\pdfoutput=1
\documentclass[11pt]{article}

\usepackage{ACL2023}

\usepackage{times}
\usepackage{latexsym}
\usepackage{booktabs}
\usepackage{array}
\usepackage{bigdelim}
\usepackage{float}
\usepackage{threeparttable}
\usepackage{subcaption}
\usepackage{marginnote}
\usepackage{enumitem}
\usepackage{awesomebox}
\usepackage{adjustbox}

\usepackage{amssymb}
\usepackage{pifont}
\newcommand{\cmark}{\ding{51}}
\newcommand{\xmark}{\ding{55}}

\usepackage[T1]{fontenc}
\usepackage[utf8]{inputenc}

\usepackage{microtype}

\usepackage{inconsolata}

\usepackage{xcolor}
\usepackage{soul}
\usepackage{colortbl}
\usepackage{graphicx}

\definecolor{snowflakeblue}{HTML}{29B5E8}
\definecolor{midblue}{HTML}{105780}
\definecolor{starblue}{HTML}{75CDD7}
\definecolor{valenciaorange}{HTML}{FF9F36}
\definecolor{firstlight}{HTML}{D45B90}
\definecolor{purplemoon}{HTML}{7254A3}
\definecolor{mediumgrey}{HTML}{5B5B5B}

\definecolor{tab1}{HTML}{4e79a7}
\definecolor{tab2}{HTML}{59a14f}
\definecolor{tab3}{HTML}{9c755f}
\definecolor{tab4}{HTML}{f28e2b}
\definecolor{tab5}{HTML}{edc948}
\definecolor{tab6}{HTML}{bab0ac}
\definecolor{tab7}{HTML}{e15759}
\definecolor{tab8}{HTML}{b07aa1}
\definecolor{tab9}{HTML}{76b7b2}
\definecolor{tab10}{HTML}{ff9da7}

\DeclareRobustCommand{\shl}[3]{\colorbox{#1}{\fontfamily{phv}\selectfont{}\small{\textcolor{#3}{#2}}}}

\DeclareRobustCommand{\gptv}{\shl{tab1}{TURBO\textsubscript{~V}}{white}}
\DeclareRobustCommand{\gptvt}{\shl{tab2}{TURBO\textsubscript{~V} + OCR}{white}}
\DeclareRobustCommand{\gptturbo}{\shl{tab3}{TURBO + OCR}{white}}
\DeclareRobustCommand{\gpt}{\shl{tab9}{8K + OCR}{white}}
\DeclareRobustCommand{\gptlong}{\shl{tab8}{32K + OCR}{white}}
\DeclareRobustCommand{\gptvreported}{\shl{tab7}{VISION}{white}}

\title{Notes on Applicability of GPT-4 to Document Understanding}

\author{Lukasz Borchmann \\
  Snowflake \\
  \texttt{lukasz.borchmann@snowflake.com} \\
}

\begin{document}

\maketitle

\begin{table*}[thb!]
   \small
   \caption{Best results achieved concerning prompt and parameters optimization (covered in Appendix~\ref{sec:vision-prompts} and \ref{sec:other-prompts}). ANLS scores except for SlideVQA, where the exact match proposed by authors is reported (both are, in fact, variants of loose accuracy). See Table~\ref{tab:sota} in the Appendix for details on the best models referenced.
   }
   \label{tab:general-results}
   \centering
   \bgroup
   \def\arraystretch{1.5}%
   % \newcolumntype{C}{ >{\raggedright\arraybackslash} m{0.79\linewidth} }
   % \newcolumntype{D}{ >{\raggedleft\arraybackslash} m{0.1\linewidth} }
   \begin{tabular}{l l l c  | c c c c}
   \toprule
   \# & Model & Version & Vision / Text & DocVQA & InfoVQA & SlideVQA & DUDE \\
   \midrule
   1 & \gptv & \texttt{1106-vision-preview} & \cmark{} / \xmark
   & $84.6$ % Local: $82.8$
   & $67.3$ % Local: $67.6$
   & $55.1$ % Local: $55.3$
   & $53.5$
   \\
   2 & \gptvt & \texttt{1106-vision-preview} & \cmark{} / \cmark
   & $87.4$
   & $71.9$
   & $57.3$ % Local: $58.2$
   & $53.3$
   \\
   % \midrule
   3 & \gptturbo & \texttt{1106-preview} & \xmark{} / \cmark 
   & $78.2$
   & $54.3$
   & $45.9$ % Local: $46.5$
   & $48.1$
   \\
   \midrule
   4 & \gpt & \texttt{0613} & \xmark{} / \cmark & $77.5$ % Local: $76.5$
   & $54.5$ % Local: $56.5$
   & $41.4$ % $42.2$
   & $48.0$ \\
   5 & \gptlong & 
   \texttt{32k-0613} & \xmark{} / \cmark & $79.5$ % Local: $77.9$
   & $52.8$ % Local: $54.4$
   & $44.7$ % $45.4$
   & $48.8$ \\
   % 0 & \texttt{[TEXT]} & $2.1$ \\
   6 & \gptvreported{} ~(reported) & & \cmark{} / \xmark  & $88.4$ & $75.1$ & --- & --- \\
   \midrule
   \multirow{2}{*}{7} & \multirow{2}{*}{Best model} & & & $93.1$ & $75.7$ & $37.7$ & $53.4$ \\
   & & & & Qwen-VL & InternVL & InstructDr & GRAM \\
   \midrule
   8 & Human performance & & & $98.1$ & $97.2$ & $89.8$ & $74.8$ \\
   \bottomrule
   \end{tabular}
   \egroup
\end{table*}

\begin{abstract}
We perform a missing, reproducible evaluation of all publicly available GPT-4 family models concerning the Document Understanding field, where it is frequently required to comprehend text spacial arrangement and visual clues in addition to textual semantics. Benchmark results indicate that though it is hard to achieve satisfactory results with text-only models, GPT-4 Vision Turbo performs well when one provides both text recognized by an external OCR engine and document images on the input. Evaluation is followed by analyses that suggest possible contamination of textual GPT-4 models and indicate the significant performance drop for lengthy documents.
\end{abstract}

\section{Introduction}

Document Understanding is the capacity to convert a document into meaningful information, which commonly involves integrating clues represented by layout, non-textual elements, and text style \cite{069059b7}. The advent of LLMs and large models able to process document images motivates the inquiry of how well they perform in this scenario compared to specialized models developed in parallel.

Though many tasks are commonly considered under the umbrella term of Document Understanding, we limit the evaluation to Document Visual Question Answering, which is the most convenient concerning both LLM-based chat assistants and the fact every piece of information within the document can be requested by expressing a question or specifying instructions in natural language.

Whereas the press release of the GPT-4 model \cite{gpt4press} mentioned scores on two document VQA datasets, the details remain unknown, and whether such results are achievable with commercially available API is vague. We intend to bridge this gap with a detailed technical report and reproducible evaluation procedure.

\paragraph{GPT-4 Family.}
As the comparison includes models trained using different recipes and architectures, each should be considered and optimized separately.
The self-reported results from March were achieved with release-day models with image inputs functionally available to a handful of selected users. Afterward, models were incrementally improved and replaced with newer variants, including the June release of newer textual GPT-4 (with 8k and 32k tokens context windows) and November releases of 'turbo' variants (capable of consuming up to 128k tokens, including tokenized images).

\begin{figure}[H]
    % \vspace{-0.6em}
    \centering
    \includegraphics[trim={20px 0 0 0},clip,width=\linewidth]{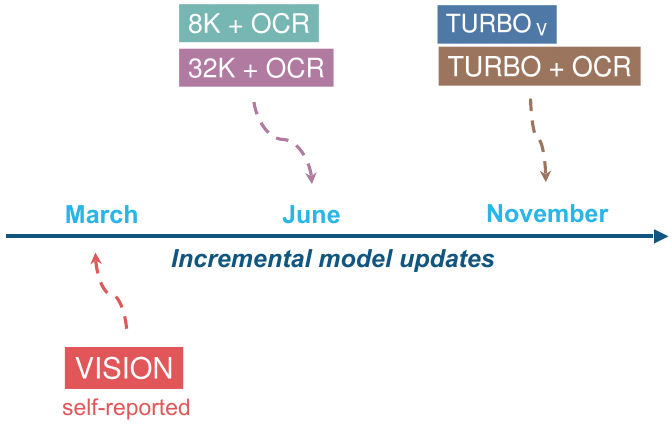}
    % \caption{Releases of GPT-4 family models.}
    \label{fig:enter-label}
    \vspace{-1.3em}
\end{figure}

\section{Experiments}

We consider DocVQA \cite{mathew2020document}, InfographicsVQA \cite{mathew2022infographicvqa}, SlideVQA \cite{SlideVQA}, and DUDE \cite{dude} datasets, as they represent the entire spectrum of document types:

\vspace{-0.2em}
\begin{table}[H]
   \label{tab:datasets}
   \centering
   \bgroup
   \small
   \setlength{\tabcolsep}{0.3em}
   \def\arraystretch{1.1}%
   \begin{tabular}{l c c c c}
   \toprule
    & DocVQA & InfoVQA & DUDE & SlideVQA \\
   \midrule
    Text-intensive & \cmark & \xmark & \cmark & \xmark \\
    Vision-intensive & \xmark & \cmark & \xmark & \cmark \\
    Multi-page & \xmark & \xmark & \cmark & \cmark \\
   \bottomrule
   \end{tabular}
   \egroup\vspace{-0.5em}
\end{table}

 % Prompts used for each model and dataset were subject to optimization described in Appendix~\ref{sec:vision-prompts} and~\ref{sec:prompts}. For setups with textual modality, we provide an input recognized by external OCR engines (Appendix~\ref{sec:ocrs}). 

\noindent This section discusses results for the best prompt, image resolution, and OCR combinations determined on validation sets (Appendixes~\ref{sec:order}-\ref{sec:other-prompts}), as well as the crucial findings of these studies.

% Table~\ref{tab:general-results}

\paragraph{Results.} Comparison in Table~\ref{tab:general-results} indicates that \gptv{} and \gptvt{} models incorporating visual aspects of the document (and, indirectly, its layout) outperform heavier text-only models.

Though we could establish state-of-the-art performance on the SlideVQA and DUDE datasets, results achieved on the well-established task of DocVQA seem poor compared to scores reported in the literature.

Importantly, we show that the GPT-4 Vision model, which can consume both text and vision modalities, benefits from providing text recognized by an external OCR engine as a part of the input. % and the document image.
Despite best-effort optimization reported in Appendixes~\ref{sec:order}-\ref{sec:other-prompts}, we could not match the undistilled GPT-4 \gptvreported{} scores in a pixel-only setup. Given the available public vision-and-text model, this level of performance seems achievable only when images are accompanied by recognized text.

% \vspace{2mm}
% \noindent Refer to appendices for detailed results and parameters optimization.

% \input{tldr}

% \paragraph{Single vs Multi-Question.}

% \paragraph{Fallback to better?} 32k for long, 8k for short. Chunk processing?

\section{Error Analysis}

We leverage the datasets' diagnostic categories and metadata to analyze models' performance depending on the input and evidence features.

\begin{table*}
   \small
   \caption{Impact of answer evidence (DocVQA and InfographicsVQA). When OCR text is provided, significant improvements in DocVQA occur with free text, forms, lists, and tables, while in InfographicsVQA, gains are most notable with visual artifacts, possibly due to the richer interplay of textual and graphic elements.}
   \label{tab:evidence-docvqa-infographics}
   \centering
   \bgroup
   \setlength{\tabcolsep}{0.63em}
   \def\arraystretch{1.5}%  1 is the default, change whatever you need
   % \newcolumntype{C}{ >{\raggedright\arraybackslash} m{0.79\linewidth} }
   % \newcolumntype{D}{ >{\raggedleft\arraybackslash} m{0.1\linewidth} }
   \begin{tabular}{l c c c c | c c c}
   \toprule
   \multirow{2}{*}{Model} & \multicolumn{4}{c|}{DocVQA} & \multicolumn{3}{c}{InfographicsVQA} \\ & Free text & Figure / Image & Form & Table / List & Textual & Figure / Visual / Map & Table / List \\
   \midrule
   \gptv & $82.8$ & $76.5$ & $88.8$ & $83.4$ & $80.0$ & $60.9$ & $67.3$ \\
   \gptvt & $\textbf{87.1}$ & $\textbf{77.4}$ & $\textbf{92.0}$ & $\textbf{87.6}$ & $\textbf{82.0}$ & $\textbf{68.4}$ & $\textbf{71.8}$ \\
   \gptturbo & $80.9$ & $55.6$ & $85.0$ & $80.5$ & $67.9$ & $43.7$ & $55.2$ \\
   \midrule
   \gpt & $80.6$ & $53.7$ & $82.7$ & $79.0$ & $65.9$ & $44.1$ & $57.2$ \\
   \gptlong & $86.2$ & $51.8$ & $85.3$ & $81.7$ & $66.9$ & $42.1$ & $57.5$ \\
   \bottomrule
   \end{tabular}
   \egroup
\end{table*}

\begin{figure*}
    \centering
    \includegraphics[width=0.65\linewidth]{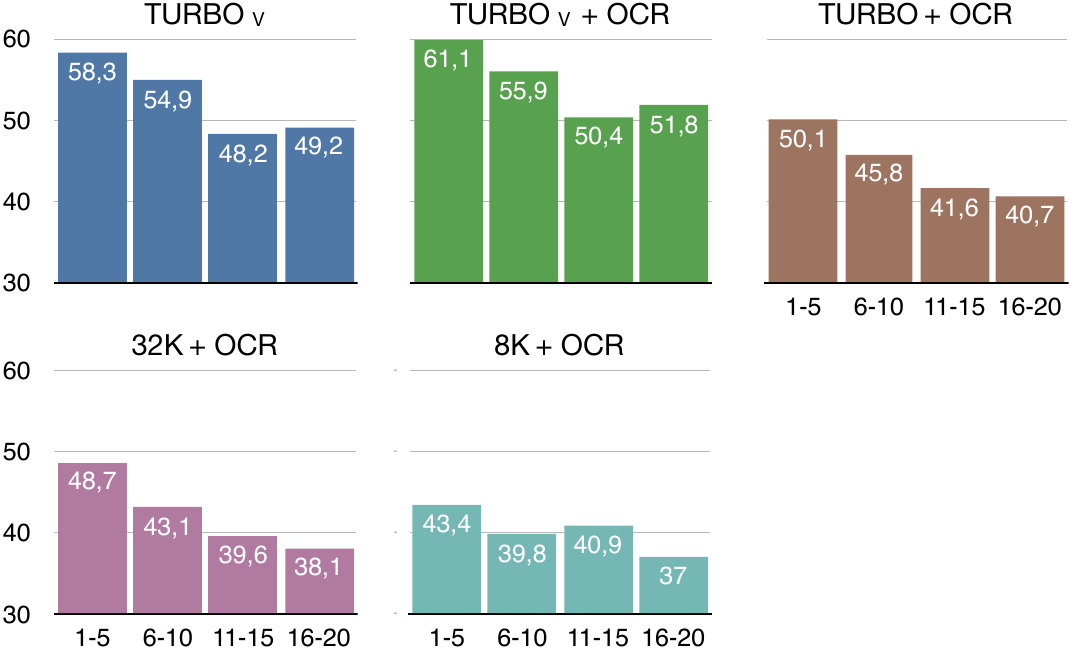}
    \caption{Scores on SlideVQA depending on the evidence location (buckets of five pages). Results reveal a \textit{primacy bias}, with higher scores when relevant information is at the beginning of the input.}
    \label{fig:evidence}
\end{figure*}

\begin{table*}
   \small
   \caption{Impact of expected answer types (DUDE). Non-turbo variants of GPT-4 demonstrate proficiency in identifying non-answerable questions despite lower overall performance stemming from a lack of visual cognition, while all models encounter challenges in generating answers for abstractive cases compared to extractive ones.}
   \label{tab:answer-type}
   \centering
   \bgroup
   \setlength{\tabcolsep}{1em}
   \def\arraystretch{1.5}%  1 is the default, change whatever you need
   % \newcolumntype{C}{ >{\raggedright\arraybackslash} m{0.79\linewidth} }
   % \newcolumntype{D}{ >{\raggedleft\arraybackslash} m{0.1\linewidth} }
   \begin{tabular}{l c c | c c}
   \toprule
   Model & Lists & Non-answerable & Abstractive & Extractive \\
   \midrule
   \gptv & $56.5$ & $57.9$ & $48.1$ & $66.1$ \\
   \gptvt & $\textbf{57.4}$ & $50.5$ & $\textbf{48.0}$ & $\textbf{68.5}$ \\
   \gptturbo & $47.8$ & $52.9$ & $30.4$ & $63.3$ \\
   \midrule
   \gpt & $42.8$ & $61.9$ & $34.6$ & $58.8$ \\
   \gptlong & $47.5$ & $\textbf{63.3}$ & $29.9$ & $63.2$ \\
   \bottomrule
   \end{tabular}
   \egroup
\end{table*}

\begin{table*}
   \small
   \caption{Impact of mentioning the dataset name (guided instruction) compared to the baseline and mentioning different datasets (misguided). Subset of 200 test sets' documents ($\sim$850 and $\sim$1,200 question-answer pairs). Change is an improvement over a maximum of baseline and misguided values.}
   \label{tab:contamination}
   \centering
   \bgroup
   \def\arraystretch{1.5}%  1 is the default, change whatever you need
   % \newcolumntype{C}{ >{\raggedright\arraybackslash} m{0.79\linewidth} }
   % \newcolumntype{D}{ >{\raggedleft\arraybackslash} m{0.1\linewidth} }
   \begin{tabular}{l c | r r r | r}
   \toprule
   Model & Dataset & Baseline & Misguided & Guided & Change \\
   \midrule
   \multirow{2}{*}{\gptv} & DocVQA & $85.3$ & $\mathbf{87.3}$ & $87.3$ & $0.0$ \\
   & InfoVQA & $68.8$ & $\mathbf{72.0}$ & $71.8$ & $-0.2$ \\
   \midrule
   \multirow{2}{*}{\gptturbo} & DocVQA & $81.1$ & $80.2$ & $\mathbf{82.8}$ & $+1.7$ \\
   & InfoVQA & $53.6$ & $58.2$ & $\mathbf{58.6}$ & $+0.4$ \\
   \midrule
   \multirow{2}{*}{\gpt} & DocVQA & $77.7$ & $78.6$ & $\mathbf{82.4}$ & $+3.8$\\
   & InfoVQA & $56.3$ & $56.0$ & $\mathbf{59.9}$ & $+3.6$ \\
   \midrule
   \multirow{2}{*}{\gptlong} & DocVQA & $82.2$ & $82.7$ & $\mathbf{85.3}$ & $+2.6$ \\
   & InfoVQA & $56.4$ & $55.5$ & $\mathbf{59.6}$ & $+3.2$ \\
   \bottomrule
   \end{tabular}
   \egroup
\end{table*}

% \begin{table*}
%    \small
%    \caption{Performance on InfographicsVQA depending on the answer evidence.}
%    \label{tab:evidence-infographics}
%    \centering
%    \bgroup
%    \def\arraystretch{1.5}%  1 is the default, change whatever you need
%    % \newcolumntype{C}{ >{\raggedright\arraybackslash} m{0.79\linewidth} }
%    % \newcolumntype{D}{ >{\raggedleft\arraybackslash} m{0.1\linewidth} }
%    \begin{tabular}{l c c c c}
%    \toprule
%    Model & Textual & Figure / Visual / Map & Table / List \\
%    \midrule
%    \gptv & $80.0$ & $60.9$ & $67.3$ \\
%    \gptvt & $\textbf{82.0}$ & $\textbf{68.4}$ & $\textbf{71.8}$ \\
%    \gptturbo & $67.9$ & $43.7$ & $55.2$ \\
%    \midrule
%    \gpt & $65.9$ & $44.1$ & $57.2$ \\
%    \gptlong & $66.9$ & $42.1$ & $57.5$ \\
%    \bottomrule
%    \end{tabular}
%    \egroup
% \end{table*}

% \begin{figure}
%     \centering
%     % \legend{0.6}{0.08}
%     \includegraphics[width=0.96\linewidth]{figures/evidence_docvqa.pdf}
%     \caption{Impact of answer evidence (DocVQA scores).}
%     \label{fig:evidence-docvqa}
%     \vspace{-10px}
% \end{figure}

\paragraph{Evidence.} Unsurprisingly, the observed advantage of \gptvt{} is echoed in scores depending on how the input document represents the information. We see that OCR-provided text significantly improves DocVQA results if the requested value is present in free text and text-rich elements structured in forms, lists, and tables (Table~\ref{tab:evidence-docvqa-infographics}). At the same time, improvement is less visible if the evidence is provided as a figure or image.

Counterintuitively, a similar analysis performed on InfographicsVQA shows that gains from OCR text presence are most apparent across visual artifacts.. We hypothesize it can be attributed to the richer interplay between textual and graphic elements, e.g., here a higher proportion of evidence requires visual and textual content to be comprehended simultaneously (36\% compared to 31\%).

% \begin{figure*}
%     \centering
%     % \legend{0.3}{0.04}\vspace{4px}
%     \includegraphics[width=0.72\linewidth]{figures/evidence_infographics.pdf}
%     \caption{Performance on InfographicsVQA depending on the answer evidence.}
%     \label{fig:evidence-infographics}
% \end{figure*}

\vspace{-0.05em}\paragraph{Evidence Location.} It has been shown that language models don't robustly use information in long input contexts \cite{liu2023lost}. Because the SlideVQA dataset provides information on the position of the requested information within a document, we can investigate how the performance of the model changes depending on the evidence location. 

Results in Figure~\ref{fig:evidence} indicate the \textit{primacy bias} for all of the models, i.e., achieved scores are highest when relevant information occurs at the beginning of the input. Moreover, they suggest \textit{recency bias} for \gptv, as the scores slightly improve as we move to the end of the input document.

\vspace{-0.05em}\paragraph{Answer Type.} As the DUDE dataset contains questions requiring a list as an answer or statements that it is non-answerable concerning the input document, we analyze the performance in these categories (Table~\ref{tab:answer-type}).

Interestingly, though list answers performance follows the general pattern, non-turbo variants of GPT-4 appear significantly better in identifying non-answerable questions --- even though their overall performance is lower due to lack of visual cognition.

When comparing extractive cases, where the answer could be copied from the input document, to abstractive cases, where it has to be generated, we observe that all models struggle with the latter scenario.

\vspace{2mm}
\noindent Concerning an operation required to provide an answer, the analysis of overall scores was not very insightful as, e.g., counting might involve counting of graphical artifacts, and comparison may require comparing visual parts of the input document. It didn't change even when we considered only answers with plain-text and tabular evidence where there was a greater chance for a fair comparison between models.

\section{Contamination Analysis}\label{sec:contamination}

As there is a possibility that train or test set splits of considered datasets were present in the training data of GPT-4, there is a question of whether it is a zero-shot performance or whether the scores were inflated due to \textit{data contamination}. The study concerns DocVQA and InfographicsVQA, as these two were published before September 2022, and it is possible that they were present in crawls used to train all of the GPT-4 models.

We use a straightforward technique of \textit{guided instruction} \cite{golchin2023time} where prompts are extended with information on the target dataset, e.g., instead of `answer the question' we use `answer the question from DocVQA dataset test split' and check if it impacts the evaluation score. Additionally, we extend the analysis with \textit{misguided instructions}, where a different than evaluated dataset name is used. This reference is helpful because mentioning even an irrelevant task could impact the performance by aligning output form with popular annotation conventions rather than end-user preferences.

Results reported in Table~\ref{tab:contamination} indicate we could increase the performance of text-only models by adding the dataset name to the prompt by up to $3.8$ points. At the same time, using a different QA dataset name does not lead to comparable outcomes for all but \gptv{} model. 
Concerning the cases when output has changed as a result of guiding and led to different scores, 40\% is a result of changing wrong or imperfect answers for answers matching the gold standard perfectly (casing included). %, e.g., from `1.44 MB' to `1.5' for the question `What is the storage capacity of 3.5" floppy disk (MB)?'

% What is the 'fund balance, end of  year' for FY1985 ?
% 109,537
% "What is the 'fund balance, end of  year' for FY1985 ?

Though the premises considered here are inconclusive, this suggests that both datasets might be present in training data of textual GPT-4 variants, and achieved scores should be taken with a grain of salt.

Though the results of the vison-enhanced model could be distinct because of the lack of contamination, it is possible it was either not exposed to dataset names (more likely in the textual crawls) or did not have an intensive to memorize answers (more substantial for textual models since some of the answers were grounded on visual layer unavailable during the training).

% This suggests that both datasets might be present in training data of textual GPT-4 variants, and achieved scores should be taken with a grain of salt. Note the premises considered here are inconclusive, and possible contamination did not require intentional action as both datasets, including their gold standards, are crawlable as a part of the RRC portal. % \footnote{\url{https://rrc.cvc.uab.es/}}

% To investigate the topic in-depth, we manually prepare redacted input documents that do not contain an answer and filter out questions that can be answered without being grounded on the inputs (e.g., resorting to general knowledge). We rerun the inference on these, expecting zero scores if the model did not memorize any parts of the test set.

% ^ Proves documents were, not annotations

% "Parity Checking" comes under which branch of Computer science?
% THEORETICAL COMPUTER SCIENCE -> INFORMATION THEORY

% 15\% of who mine data for patterns
% 15\% -> Data scientists

% Among the paid software and web-based finance programs, which two are the cheapest given?
% InEx Finance, BUDGT
% InEx Finance, SPENDEE
 % "who was the president in the year 2000?", "values": "value": "Bill Clinton" -> Not mentioned

% \section{Pricing}

\section{Limitations}

It is important to acknowledge certain limitations that shape the context of analysis.

\paragraph{Only performance.} Our analysis is limited to the performance of the models, as expressed by their scores on popular benchmark datasets. There are, however, other vital aspects to consider before deploying models, such as data privacy, cost, legal compliance, or biases that these can exhibit.

\paragraph{Dataset selection.} Though we sketched the rationale behind the selection of datasets for models' comparison, it is debatable. Different choices of datasets could lead to different outcomes. Moreover, the selection impacted results indirectly, e.g., we were limited to the source image sizes, sometimes leading to upsampling for higher-resolution configurations (it is possible that with control over the digitization, one can achieve better results, despite the fact our results indicate no gain from resolution above 2k pixels).

\paragraph{No finetuning.} Unlike most state-of-the-art references, our evaluation procedure did not assume finetuning on a train set of considered datasets. Though some of these data could unintentionally end up in the GPT-4 training mix (see Section~\ref{sec:contamination}), finetuned models of this size could, in principle, vastly improve their performance. % Nevertheless, the possibility of performing such an action in closed models is minimal.

\paragraph{Vast search space.} Zero-shot performance of LLMs depends heavily on used prompts. Since these can be tested and refined based merely on the author's intuition that does not obey obvious and unambiguous strategies, one never knows if another prompt could lead to significantly higher scores. Moreover, all optimizations were performed on a subset of the validation set to reduce the cost of experiments. It is possible that testing all combinations of parameters on a complete dev set would lead to other choices.

\paragraph{Huge impact of OCR.} The performance of text-only models depends on how the used OCR engine preserved and represented the local structure of the document. Since even the best model cannot achieve perfect scores with imperfect inputs, it is possible to achieve better results with different input representations or better textual content recognition.

\paragraph{Third-party dependency.} As models we consider are accessible only via APIs we have no control over, some inevitable or hypothetical limits are involved. First, it happened infrequently that the question-document pair was rejected because of the provider's content filtering policy, slightly decreasing the overall score. Second, we treated OpenAI and Azure APIs with the same model interchangeably, assuming that if there are some differences, they are negligible. Finally, we cannot guarantee a lack of behavioral changes during the experiment.

\paragraph{Specific 'understanding.'} When document processing systems are considered, the term 'understanding' we use has a very narrow and specific meaning. It does not imply the ability of models to reason or their intelligence.

\vspace{2mm}
\noindent These limitations provide important context and present opportunities for future research to address.

\section{Future Considerations}

We anticipate a few potential areas for forthcoming studies, including ways to improve answer quality and dimensions to measure performance across.

\paragraph{Text arrangement.} Preserving text reading order and its integrity, as well as comprehension of spatial arrangement, are all crucial for Document Understanding problems. These can be manipulated by OCR setup (e.g., Tesseract can represent text arrangement with spaces, and Azure to produce the human-friendly reading order) or represented as a part of plain text input. Studying the impact of different input text representations and arrangements might be worthwhile, particularly when considering plain-text LLMs.

\paragraph{Confidence calibration.} Document Understanding systems are commonly considered in business process automation, where it is crucial to have well-calibrated confidence scores. Since the API of considered models does not provide a single confidence score for a generated answer, one is limited to methods based on, e.g., drawing several answers from the model, per-token probabilities, or the \textit{verbalized} confidence \cite{lin2022teaching}. It is worth evaluating such estimates' reliability. % in a separate study.

\paragraph{Multi-QA.} As datasets under consideration commonly had a few questions posed to the same input document, it would be cost-efficient to formulate prompts to extract multiple values simultaneously. Moreover, though the impact of such an approach remains unknown, it can potentially increase the answer's quality \cite{dwojak-etal-2020-dataset}.

\vspace{2mm}
\noindent We eagerly look forward to further exploration in these areas.

\section{Summary}

We assessed the performance of GPT-4 family models in document understanding, extending previous evaluations by including the DUDE and SlideVQA datasets. % Thus, unlike prior studies, which focused solely on single-page documents, our analysis covers a broader spectrum. Importantly, in contrast to the undisclosed setup details in previous reports, we provide optimized prompts, OCR, and input image resolutions, ensuring reproducibility with publicly available APIs. 
With a detailed analysis of the results and commentary, we provided practical and theoretical insights into the field, in particular:
\begin{itemize}[leftmargin=*]
    \item demonstrated that even models of this parameter count underperform in image-only setup and vastly benefit from providing text in addition to the input image;
    \item revealed a \textit{primacy bias} in model performance, e.g., the tendency to perform significantly better with requested information at the beginning of input documents;
    \item proposed an extension to the \textit{guided instruction} contamination assessing technique;
    \item identified categories of questions compared models underperform with and suggested prompts, OCRs, and input image resolutions can be used to improve their performance.
\end{itemize}

\noindent Importantly, in contrast to the undisclosed setup in previous reports, we provide all the details, ensuring reproducibility with publicly available APIs. 

% The study acknowledges limitations and suggests future research directions, including multi-question answering and confidence calibration for document understanding systems.

% \clearpage
\bibliography{anthology,custom}
\bibliographystyle{acl_natbib}

\clearpage
\appendix

\begin{figure*}[bht!]
    \centering
    \includegraphics[width=0.95\textwidth]{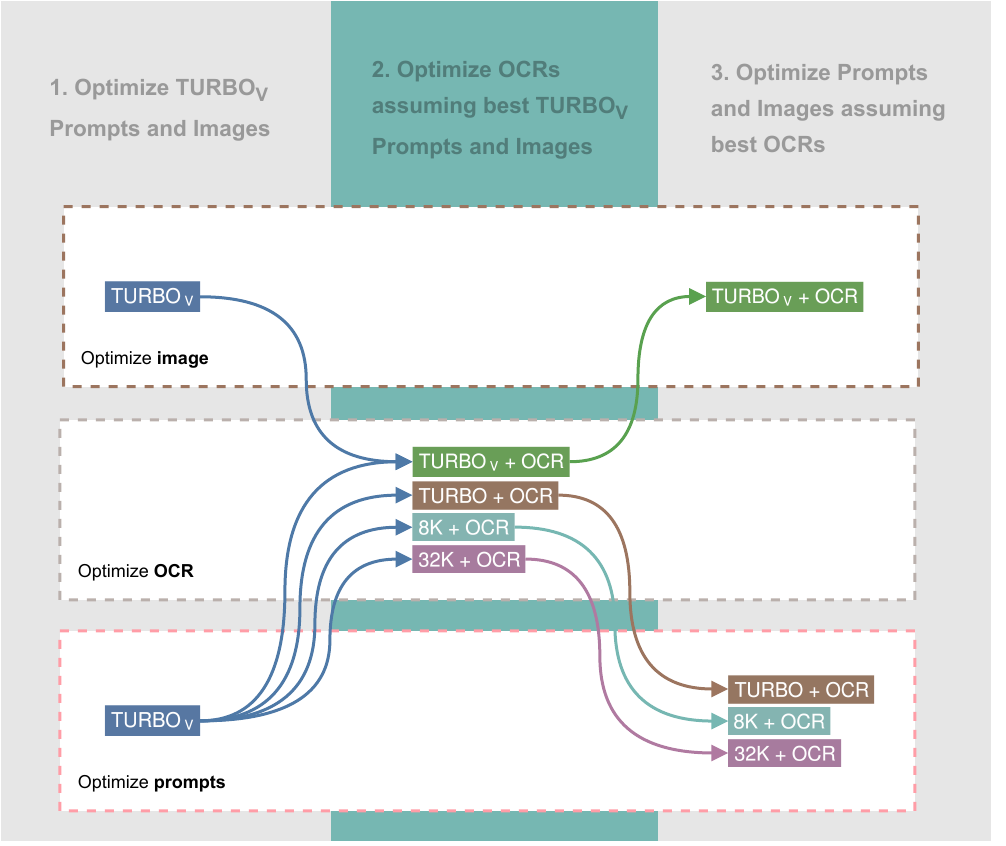}
    \caption{The order of experiments assumed to reduce the search space and not optimize all combinations of prompts and inputs (image properties, OCRs).}
    \label{fig:order}
\end{figure*}

\section{Order of Experiments}\label{sec:order}

For practical reasons, we did not optimize all combinations of prompts and inputs (image properties, OCRs). Instead, experiments are performed in a chosen order, which reduces the search space (Figure~\ref{fig:order}).

We start by optimizing prompts on \gptv{} assuming original resolution (SlideVQA) or high-resolution images (2048px on the longer side for the rest of the datasets considered).

The best \gptv{} prompts are assumed during the comparison of OCR engines for the remaining models, as well as input image format and resolution for \gptv{}.

Next, we optimize prompts for the rest of the models, assuming the OCRs they perform best with. Finally, we calculate results in Table~\ref{tab:general-results}, taking each model's best prompts, OCRs, and input image sizes.

\section{Optimizing Prompt for Vision Model}\label{sec:vision-prompts}

\paragraph{DocVQA, InfographicsVQA, SlideVQA.} We evaluate the impact of different choices on a subset of 50 documents from the validation set of each dataset, i.e., approx. 200, 400, and 300 question-answer pairs.

Considered variants, shown in Table~\ref{tab:prompt-gpt4v-docvqa}, result from the fact the model (1) tends to produce boilerplate text in addition to the value expected concerning the convention used in the dataset, (2) in some cases, the model refrains from answering, whereas all of the questions in both datasets can be answered based on the document provided.

Results indicate that Prompt~5 performs robustly across all datasets and is topped only on the DocVQA dataset by a negligible margin.

\begin{table*}[h]
   \small
   \caption{Impact of different \gptv{}  prompts on DocVQA, InfographicsVQA, and SlideVQA validation scores.} 
   \label{tab:prompt-gpt4v-docvqa}
   \centering
   \bgroup
   \def\arraystretch{1.5}%  1 is the default, change whatever you need
   % \newcolumntype{C}{ >{\raggedright\arraybackslash} m{0.79\linewidth} }
   % \newcolumntype{D}{ >{\raggedleft\arraybackslash} m{0.1\linewidth} }
   \begin{tabular}{c p{0.58\linewidth} r r r}
   \toprule
   \# & Prompt & DocVQA & InfoVQA & SlideVQA \\ 
   \midrule
   % 0 & \texttt{[TEXT]} & $2.1$ \\
   1 & Answer the question: \texttt{[TEXT]} & $2.1$ & $0.0$ & $0.0$ \\
   2 & Answer the question. Do not write a full sentence, just provide a value. Question: \texttt{[TEXT]}  & $\mathbf{87.5}$ & $67.1$ & $\textbf{63.1}$ \\
   3 & Answer the question. Be concise and provide a value I~am looking for only. Question: \texttt{[TEXT]} & $80.4$ & $57.5$ & $48.1$ \\
   4 & Answer the question. Do not write a full sentence, just provide a value. Always try to provide an answer. Question: \texttt{[TEXT]} & $85.0$ & $67.1$ & $61.4$ \\
   5 & Answer the question. Do not write a full sentence, just provide a value. If the value is unclear, guess it given an input document. Question: \texttt{[TEXT]} & $87.4$ & $\mathbf{68.7}$ & $62.4$ \\
   % 6 & Consider the question: \texttt{[TEXT]}. Based on the context image, the answer would be: & \\
   6 & Replace \texttt{[ANSWER]} with a value in the template given question and document. $\hookleftarrow$ Question: \texttt{[TEXT]} $\hookleftarrow$  Template: Based on the context, the answer to the question would be "\texttt{[ANSWER]}". & $85.1$ & $61.8$ & $59.7$ \\
   \bottomrule
   \end{tabular}
   \egroup
   
\end{table*}

\begin{table*}
   \small
   \caption{Impact of different \gptv{}  prompts on DUDE validation scores.} 
   \label{tab:prompt-gpt4v-dude}
   \centering
   \bgroup
   \def\arraystretch{1.5}%  1 is the default, change whatever you need
   % \newcolumntype{C}{ >{\raggedright\arraybackslash} m{0.79\linewidth} }
   % \newcolumntype{D}{ >{\raggedleft\arraybackslash} m{0.1\linewidth} }
   \begin{tabular}{c p{0.81\linewidth} r}
   \toprule
   \# & Prompt & DUDE \\ 
   \midrule
   5 & Answer the question. Do not write a full sentence, just provide a value. If the value is unclear, guess it given an input document. Question: \texttt{[TEXT]} & $38.6$ \\
   % 0 & \texttt{[TEXT]} & $2.1$ \\
   7 & Answer the question. Do not write a full sentence. Provide a value as a Python list. If there is a single answer, the output should be a one-element list like ["ANSWER"]. If there are multiple valid answers, the list will have several elements, e.g., ["ANSWER 1", "ANSWER 2"]. The output should be ["None"] if the value is unclear. Question: \texttt{[TEXT]} & $40.1$ \\
   8 & Answer the question. Do not write a full sentence. Provide a value as a Python list. If there is a single answer, the output should be a one-element list like ["ANSWER"]. If there are multiple valid answers, the list will have several elements, e.g., ["ANSWER 1", "ANSWER 2"]. The output should be ["None"] if the value cannot be found in the document. Question: \texttt{[TEXT]} & $\mathbf{42.7}$ \\
   9 & Provide a value as a Python list. If there is a single answer, the output should be a one-element list like ["John"]. If there are multiple valid answers, the list will have several elements, e.g., ["1997", "1998"]. The output should be ["None"] if the value cannot be found in the document. The values we are looking for are related to the question: \texttt{[TEXT]} & $41.6$ \\
   10 & Answer the question: \texttt{[TEXT]} $\hookleftarrow$ Do not write a full sentence. Provide a value as a Python list, e.g., ["ANSWER"]. If there are multiple valid answers, the list will have several elements, e.g., ["ANSWER 1", "ANSWER 2"]. The output should be ["None"] if the value cannot be found in the document. & $39.7$ \\
    11 & Answer the question: \texttt{[TEXT]} $\hookleftarrow$  Keep the answer short. Respond "None" if not sure about the answer. If there are multiple valid answers, separate them by "|." & $34.0$ \\
    12 & Answer the question: \texttt{[TEXT]} $\hookleftarrow$ Do not write a full sentence, just provide a value. Respond "None" if the value cannot be found in the document. If there are multiple valid values, separate them by "|."  & $34.4$ \\
   \bottomrule
   \end{tabular}
   \egroup
\end{table*}

\paragraph{DUDE.} As DUDE introduces novel types of answers, we perform prompt optimization for this dataset separately. We start with the most robust prompt from Table~\ref{tab:prompt-gpt4v-docvqa} and consider variants (Table~\ref{tab:prompt-gpt4v-dude}) attempted to cover list values and statements that the question cannot be answered grounded on the input document. We evaluate the impact of different choices on a subset of 30 documents from the validation set, i.e., 300 question-answer pairs.

\section{Optimizing Image Resolution}

We optimize the input image, assuming the same subsets of validation sets as in the previous section.

Since the \gptv{} consumes a low-resolution representation of the entire input image and, optionally, high-resolution crops, we investigate the impact of high-resolution crops' presence and underlying image size.

Results in Table~\ref{tab:images-v} indicate that the high-resolution variant with the image size of $2048$px on the longer side performs robustly both with and without OCR output and independently from the dataset. It was outperformed only once by a small margin with twice the lower resolution.

This observation holds for \gptvt{} (Table~\ref{tab:images-vt}), where one can additionally observe that the gain from text availability is vast in a low-resolution regime.

\section{Optimizing OCRs}\label{sec:ocrs}

Analogously to the previous studies, we evaluate Tesseract 5.3.3, Azure Cognitive Services 3.2 (2022-04-30), and Amazon Textract (Detect Document Text 1.0) OCRs. Experiments in this section assume Prompt~8 for DUDE, Prompt~2 for DocVQA, and Prompt~5 for the rest of the datasets considered.

Though there is no universally superior solution, Table~\ref{tab:ocrs} indicates that Textract is the best choice and the Tesseract is the worst for most datasets, assuming the text-only regime. When accompanied by vision, the results become noisier, and the model seems to benefit even from the availability of low-quality OCR.

Note that this study assumed each model's default settings, and it is possible that one can achieve different rankings of OCRs when manipulating, e.g., reading order parameters. Nevertheless, we expect the default parameter values to be optimized by the providers, and fine-grained OCR optimization is outside the scope of this paper.

\section{Optimizing Prompt for Text Models}\label{sec:other-prompts}

Having the OCRs, we compare how prompts from Section~\ref{sec:vision-prompts} behave on text-only models. This leads to several changes, which is not unexpected concerning zero-shot setup, which is extremely sensitive to prompt choice, and fact prompts generally do not transfer between different models, even if they retain some architectural and training data similarities.

This concludes the parameter optimization, leading to the final setup in Table~\ref{tab:final-setup}.

% \section{Pricing}

\begin{table*}
   \small
   \caption{Study of Tesseract 5.3.3, Azure Cognitive Services 3.2 (2022-04-30), and Amazon Textract (Detect Document Text 1.0). Prompt~2 (DocVQA, SlideVQA), Prompt~5 (InfographicsVQA), and Prompt~8 (DUDE).}
   \label{tab:ocrs}
   \centering
   \bgroup
   \def\arraystretch{1.5}%
   % \newcolumntype{C}{ >{\raggedright\arraybackslash} m{0.79\linewidth} }
   % \newcolumntype{D}{ >{\raggedleft\arraybackslash} m{0.1\linewidth} }
   \begin{tabular}{l l r r r r}
   \toprule
   Model & OCR & DocVQA & InfoVQA & SlideVQA & DUDE \\
   \midrule
   \multirow{3}{*}{\gptvt} & Tesseract & $87.4$ & $\mathbf{72.7}$ & $62.4$ & $43.9$ \\
   & Azure & $\mathbf{87.5}$ & $71.0$ & $62.7$ & $\mathbf{45.2}$ \\
   & Amazon & $87.1$ & $72.4$ & $\mathbf{63.1}$ & $44.3$ \\
   \midrule
   \multirow{3}{*}{\gptturbo} & Tesseract & $65.6$ & $41.1$ & $43.4$ & $34.5$ \\
   & Azure & $79.4$ & $56.4$ & $\mathbf{54.6}$ & $\mathbf{38.3}$ \\
   & Amazon & $\mathbf{82.1}$ & $\mathbf{58.9}$ & $51.2$ & $37.6$ \\
   \midrule
   \multirow{3}{*}{\gpt} & Tesseract & $64.6$ & $40.5$ & $36.9$ & $31.3$ \\
   & Azure & $81.7$ & $57.0$ & $44.7$ & $\mathbf{36.7}$ \\
   & Amazon & $\mathbf{82.6}$ & $\mathbf{59.1}$ & $\textbf{45.8}$ & $34.2$ \\
   \midrule
   \multirow{3}{*}{\gptlong} & Tesseract & $64.2$ & $34.3$ & $41.0$ & $37.4$ \\
   & Azure & $76.4$ & $53.1$ & $51.9$ & $\mathbf{39.3}$ \\
   & Amazon & $\mathbf{76.9}$ & $\mathbf{56.4}$ & $\mathbf{52.9}$ & $39.2$ \\
   \bottomrule
   \end{tabular}
   \egroup
\end{table*}

\begin{table*}
   \small
   \caption{Impact of different prompts on DocVQA, InfographicsVQA, and SlideVQA validation scores assuming the best OCR for each dataset (selected in the ablation study).} 
   \label{tab:prompt-other}
   \centering
   \bgroup
   \def\arraystretch{1.5}%  1 is the default, change whatever you need
   % \newcolumntype{C}{ >{\raggedright\arraybackslash} m{0.79\linewidth} }
   % \newcolumntype{D}{ >{\raggedleft\arraybackslash} m{0.1\linewidth} }
   \begin{tabular}{l l r r r r r r}
   \toprule
   Model & Dataset & Prompt 1 & Prompt 2 & Prompt 3 & Prompt 4 & Prompt 5 & Prompt 6 \\
   \midrule
   % & DocVQA & & & & & \\
   % \gptvt{} & InfographicsVQA & & & & & \\
   % & SlideVQA & & & & & \\
   % \midrule
   & DocVQA & $1.0$ & $82.1$ & $67.0$ & $78.7$ & $80.6$ & $\mathbf{84.7}$ \\
   \gptturbo{} & InfographicsVQA & $0.0$ & $58.2$ & $43.8$ & $\mathbf{59.4}$ & $58.9$ & $58.7$ \\
   & SlideVQA & $0.0$ & $54.6$ & $44.7$ & $52.5$ & $54.2$ & $\mathbf{54.9}$ \\
   \midrule
   & DocVQA & $3.4$ & $\mathbf{82.6}$ & $16.7$ & $79.0$ & $80.1$ & $80.6$ \\
   \gpt{} & InfographicsVQA & $1.2$ & $57.1$ & $9.7$ & $56.8$ & $59.1$ & $\mathbf{59.5}$ \\
   & SlideVQA & $0.0$ & $45.8$ & $1.4$ & $44.1$ & $45.1$ & $\mathbf{46.8}$ \\
   \midrule
   & DocVQA & $3.0$ & $76.9$ & $8.2$ & $76.8$ & $77.8$ & $\mathbf{80.4}$ \\
   \gptlong{} & InfographicsVQA & $0.1$ & $58.2$ & $14.6$ & $\mathbf{58.4}$ & $56.4$ & $58.2$ \\
   & SlideVQA & $0.0$ & $\mathbf{52.9}$ & $11.2$ & $49.2$ & $48.8$ & $52.5$ \\
   \bottomrule
   \end{tabular}
   \egroup
\end{table*}

%    & & & & QwenVL & InternVL & InstructDr & GRAM \\

\begin{table*}
   \centering
   \bgroup
   \small
   \begin{threeparttable}
   \caption{Sources of human performance and assumed state-of-the-art models.} \label{tab:sota}
   \begin{tabular}{l l l}
   \toprule
    Dataset & State-of-the-art & Human performance \\
   \midrule
    DocVQA & Qwen-VL-Max \cite{Qwen-VL} & \citet{mathew2020document} \\
    InfographicsVQA & InternVL \cite{chen2023internvl,chen2024far} & \citet{mathew2022infographicvqa} \\
    SlideVQA & InstructDr \cite{tanaka2024instructdoc} & \citet{SlideVQA} \\
    DUDE & GRAM \cite{blau2024gram} & \citet{dude} \\
   \bottomrule
   \end{tabular}
   % \vspace{0.5em}
   %  \begin{tablenotes}
   %  \scriptsize
   %  \item[1] \url{https://rrc.cvc.uab.es/?ch=17&com=evaluation&view=method_info&task=1&m=99110}
   %  \item[2] \url{https://rrc.cvc.uab.es/?ch=17&com=evaluation&view=method_info&task=3&m=105897}
   %  \item[3] \url{https://rrc.cvc.uab.es/?ch=23&com=evaluation&view=method_info&task=1&m=104106}
   %  \end{tablenotes}
   \end{threeparttable}
   \egroup
\end{table*}

\begin{table*}
   \small
   \caption{Impact of different prompts on DUDE validation scores assuming the best OCR for each dataset (selected in the ablation study).} 
   \label{tab:prompt-other-dude}
   \centering
   \bgroup
   \def\arraystretch{1.7}%  1 is the default, change whatever you need
   % \newcolumntype{C}{ >{\raggedright\arraybackslash} m{0.79\linewidth} }
   % \newcolumntype{D}{ >{\raggedleft\arraybackslash} m{0.1\linewidth} }
   \begin{tabular}{l r r r r r r r}
   \toprule
   Model & Prompt 5 & Prompt 7 & Prompt 8 & Prompt 9 & Prompt 10 & Prompt 11 & Prompt 12 \\ 
   \midrule
   % \gptvt{} & \\
   \gptturbo{} & $34.2$ & $37.9$ & $38.3$ & $\mathbf{38.8}$ & $36.6$ & $34.0$ & $30.6$ \\
   \gpt{} & $31.8$ & $32.6$ & $36.7$ & $36.0$ & $36.4$ & $26.0$ & $\mathbf{40.2}$  \\
   \gptlong{} & $33.1$ & $36.0$ & $\mathbf{39.3}$ & $34.8$ & $34.5$ & $33.9$ & $33.5$ \\
   \bottomrule
   \end{tabular}
   \egroup
\end{table*}

\begin{table}
   \small
   \caption{Final test set evaluation setup optimized on validation sets.} 
   \label{tab:final-setup}
   \centering
   \bgroup
   \def\arraystretch{1.7}%  1 is the default, change whatever you need
   % \newcolumntype{C}{ >{\raggedright\arraybackslash} m{0.79\linewidth} }
   % \newcolumntype{D}{ >{\raggedleft\arraybackslash} m{0.1\linewidth} }
   \begin{tabular}{l c c c c}
   \toprule
   Dataset & Prompt & OCR & \multicolumn{2}{c}{Image} \\ 
   \midrule
   \multicolumn{5}{l}{\gptv{}} \\
   DocVQA & 2 & --- & $2048$px & jpg \\
   InfoVQA & 5 & --- & $2048$px & png \\
   SlideVQA & 2 & --- & $2048$px & jpg \\
   DUDE & 8 & --- & $1024$px & jpg \\
   \midrule
   \multicolumn{5}{l}{\gptvt{}} \\
   DocVQA & 2 & Azure & $2048$px & jpg \\
   InfoVQA & 5 & Tesseract & $2048$px & png \\
   SlideVQA & 2 & Amazon & $1024$px & jpg \\
   DUDE & 8 & Azure & $2048$px & jpg \\
   \midrule
   \multicolumn{5}{l}{\gptturbo{}} \\
   DocVQA & 6 & Amazon & \multicolumn{2}{c}{---} \\
   InfoVQA & 4 & Amazon & \multicolumn{2}{c}{---} \\
   SlideVQA & 6 & Azure & \multicolumn{2}{c}{---}  \\
   DUDE & 9 & Azure & \multicolumn{2}{c}{---} \\
   \midrule
   \multicolumn{5}{l}{\gpt{}} \\
   DocVQA & 2 & Amazon & \multicolumn{2}{c}{---} \\
   InfoVQA & 6 & Amazon & \multicolumn{2}{c}{---} \\
   SlideVQA & 6 & Amazon & \multicolumn{2}{c}{---} \\
   DUDE & 12 & Azure & \multicolumn{2}{c}{---} \\
   \midrule
   \multicolumn{5}{l}{\gptlong{}} \\
   DocVQA & 6 & Amazon  & \multicolumn{2}{c}{---} \\
   InfoVQA & 4 & Amazon  & \multicolumn{2}{c}{---} \\
   SlideVQA & 2 & Amazon  & \multicolumn{2}{c}{---} \\
   DUDE & 8 & Azure  & \multicolumn{2}{c}{---} \\
   \bottomrule
   \end{tabular}
   \egroup
\end{table}

\begin{table}
       \caption{Impact of different image sizes on scores. Pixels on the longer side except for \textit{low} that corresponds to \textit{detail: low} in the API (512px square).} 
   \begin{subtable}{\linewidth}
   \small
   \caption{\gptv} 
   \label{tab:images-v}
   \centering
   \bgroup
   \def\arraystretch{1.5}%  1 is the default, change whatever you need
   % \newcolumntype{C}{ >{\raggedright\arraybackslash} m{0.79\linewidth} }
   % \newcolumntype{D}{ >{\raggedleft\arraybackslash} m{0.1\linewidth} }
   \begin{tabular}{l r r r r r}
   \toprule
   Dataset & low & 512 & 1024 & 2048 & 4096 \\
   \midrule
   % & DocVQA & & & & & \\
   % \gptvt{} & InfographicsVQA & & & & & \\
   % & SlideVQA & & & & & \\
   % \midrule
   DocVQA & $50.5$ & $69.3$ & $85.2$ & $\mathbf{87.5}$ & $87.2$ \\
   InfoVQA & $37.8$ & $45.2$ & $58.8$ & $\textbf{68.7}$ & $68.0$ \\
   SlideVQA & $47.8$ & $62.7$ & $63.1$ & $\textbf{64.7}$ & $61.0$ \\
   DUDE & $30.3$ & $34.1$ & $\textbf{44.2}$ & $43.8$ & $44.0$ \\
   \bottomrule
   \end{tabular}
   \egroup
   \vspace{3mm}
   \end{subtable}
   \begin{subtable}{\linewidth}
       \small
       \caption{\gptvt} 
       \label{tab:images-vt}
       \centering
       \bgroup
       \def\arraystretch{1.5}%  1 is the default, change whatever you need
       % \newcolumntype{C}{ >{\raggedright\arraybackslash} m{0.79\linewidth} }
       % \newcolumntype{D}{ >{\raggedleft\arraybackslash} m{0.1\linewidth} }
       \begin{tabular}{l r r r r r}
       \toprule
       Dataset & low & 512 & 1024 & 2048 & 4096 \\
       \midrule
       % & DocVQA & & & & & \\
       % \gptvt{} & InfographicsVQA & & & & & \\
       % & SlideVQA & & & & & \\
       % \midrule
       DocVQA & $80.6$ & $86.4$ & $86.5$ & $\mathbf{87.5}$ & $87.1$ \\
       InfoVQA & $52.9$ & $57.7$ & $63.1$ & $\mathbf{72.7}$ & $72.6$ \\
       SlideVQA & $54.2$ & $63.7$ & $\mathbf{64.7}$ & $63.1$ & $62.0$ \\
       DUDE & $35.2$ & $35.2$ & $44.6$ & $\mathbf{45.2}$ & $43.8$ \\
       \bottomrule
       \end{tabular}
       \egroup
   \end{subtable}
\end{table}

\begin{table}
   \small
   \caption{Guided (G) and misguided (M) prompts used in the Contamination Analysis.} 
   \label{tab:guided-prompts}
   \centering
   \bgroup
   \def\arraystretch{1.5}%  1 is the default, change whatever you need
   % \newcolumntype{C}{ >{\raggedright\arraybackslash} m{0.79\linewidth} }
   % \newcolumntype{D}{ >{\raggedleft\arraybackslash} m{0.1\linewidth} }
   \begin{tabular}{c p{0.79\linewidth}}
   \toprule
    % \# & Prompt \\
    % \midrule
    2G & Answer the question \textbf{from DocVQA test split (Task 1 - Single Page Document VQA)}. Do not write a full sentence, just provide a value. Question: \texttt{[TEXT]} \\
    2M & Answer the question \textbf{from SQuAD test split}. Do not write a full sentence, just provide a value. Question: \texttt{[TEXT]} \\
    4G & Answer the question \textbf{from Infographics VQA test split}. Do not write a full sentence, just provide a value. Always try to provide an answer. Question: \texttt{[TEXT]} \\
    4M & Answer the question \textbf{from SQuAD test split}. Do not write a full sentence, just provide a value. Always try to provide an answer. Question: \texttt{[TEXT]} \\
    6G & Replace \texttt{[ANSWER]} with a value in the template given question and document. $\hookleftarrow$ \textbf{Source: DocVQA test split (Task 1 - Single Page Document VQA).} $\hookleftarrow$ Question: \texttt{[TEXT]} $\hookleftarrow$  Template: Based on the context, the answer to the question would be "\texttt{[ANSWER]}". \\
     & Replace \texttt{[ANSWER]} with a value in the template given question and document. $\hookleftarrow$ \textbf{Source: Infographics VQA test split.} $\hookleftarrow$ Question: \texttt{[TEXT]} $\hookleftarrow$  Template: Based on the context, the answer to the question would be "\texttt{[ANSWER]}". \\
    6M & Replace \texttt{[ANSWER]} with a value in the template given question and document. $\hookleftarrow$ \textbf{Source: SQuAD test split.} $\hookleftarrow$ Question: \texttt{[TEXT]} $\hookleftarrow$  Template: Based on the context, the answer to the question would be "\texttt{[ANSWER]}". \\

   \bottomrule
   \end{tabular}
   \egroup
\end{table}

\end{document}